\documentclass{article}

\PassOptionsToPackage{numbers}{natbib}

\usepackage[final]{nips_2017}

\usepackage[utf8]{inputenc} % allow utf-8 input
\usepackage[T1]{fontenc}    % use 8-bit T1 fonts
\usepackage{hyperref}       % hyperlinks
\usepackage{url}            % simple URL typesetting
\usepackage{booktabs}       % professional-quality tables
\usepackage{amsfonts}       % blackboard math symbols
\usepackage{nicefrac}       % compact symbols for 1/2, etc.
\usepackage{microtype}      % microtypography
\usepackage{graphicx}       % png support
\usepackage[frozencache]{minted}         % for code listings
\usepackage{caption}

\newenvironment{longlisting}{\captionsetup{type=listing}}{}

\title{Horovod: fast and easy distributed deep learning in TensorFlow}

\author{
  Alexander Sergeev\\
  Uber Technologies, Inc.\\
  \texttt{asergeev@uber.com} \\
  \And
  Mike Del Balso\\
  Uber Technologies, Inc.\\
  \texttt{mdb@uber.com} \\
}

\begin{document}

\maketitle

\begin{abstract}
  Training modern deep learning models requires large amounts of
  computation, often provided by GPUs. Scaling computation from one GPU
  to many can enable much faster training and research progress but
  entails two complications. First, the training library must support
  inter-GPU communication. Depending on the particular methods employed,
  this communication may entail anywhere from negligible to significant
  overhead. Second, the user must modify his or her training code to take
  advantage of inter-GPU communication. Depending on the training
  library's API, the modification required may be either significant or
  minimal.

  Existing methods for enabling multi-GPU training under the TensorFlow
  library entail non-negligible communication overhead and require users
  to heavily modify their model-building code, leading many researchers
  to avoid the whole mess and stick with slower single-GPU training.
  In this paper we introduce Horovod, an open source library that improves
  on both obstructions to scaling: it employs efficient inter-GPU
  communication via ring reduction and requires only a few lines of
  modification to user code, enabling faster, easier distributed training
  in TensorFlow. Horovod is available under the Apache 2.0 license at
  \mbox{\url{https://github.com/uber/horovod}}.
\end{abstract}

\section{Introduction}

Over the past few years, advances in deep learning have driven
tremendous progress in image processing, speech recognition, and
forecasting. At Uber, we apply deep learning across our business;
from self-driving research to trip forecasting and fraud prevention,
deep learning enables our engineers and data scientists to create
better experiences for our users.

TensorFlow~\cite{tensorflow} has become a preferred deep learning
library at Uber for a variety of reasons. To start, the framework
is one of the most widely used open source frameworks for deep learning,
which makes it easy to onboard new users. It also combines high
performance with an ability to tinker with low-level model details---for
instance, we can use both high-level APIs, such as Keras~\cite{keras},
and implement our own custom operators using NVIDIA's CUDA toolkit.
Additionally, TensorFlow has end-to-end support for a wide variety
of deep learning use cases, from conducting exploratory research to
deploying models in production on cloud servers, mobile apps, and
even self-driving vehicles.

In September 2017, Uber Engineering introduced Michelangelo~\cite{michelangelo},
an internal ML-as-a-service platform that democratizes machine learning
and makes it easy to build and deploy these systems at scale. In this paper,
we introduce Horovod, an open-source component of Michelangelo's deep learning
toolkit which makes it easier to start---and speed up---distributed deep
learning projects with TensorFlow.  Horovod is available under the Apache 2.0
license at~\url{https://github.com/uber/horovod}.

\section{Going distributed}

As we began training more and more machine learning models at Uber,
their size and data consumption grew significantly. In a large portion
of cases, the models were still small enough to fit on one or multiple
GPUs within a server, but as datasets grew, so did the training times,
which sometimes took a week or longer to complete. We found ourselves
in need of a way to train using a lot of data while maintaining short
training times. To achieve this, our team turned to distributed training.

We began by testing the standard distributed TensorFlow~\cite{tensorflow_ps}
technique. After trying it out on a few models, it became apparent that we
needed to make two adjustments.

First, after following the documentation and code examples, it was not
always clear which code modifications needed to be made to distribute
their model training code. The standard distributed TensorFlow package
introduces many new concepts: workers, parameter servers, \texttt{tf.Server()},
\texttt{tf.ClusterSpec()}, \texttt{tf.train.SyncReplicasOptimizer()}, and
\texttt{tf.train.replicas\_device\_setter()} to name a few. 

While this API may be well-suited to certain scenarios, in many cases it
introduced subtle, hard-to-diagnose bugs. Identifying and fixing these
bugs unfortunately required users to climb a steep learning curve of
concepts they almost never care about---they just want to take an existing
model and make it faster, not become an expert along the way in syncronization
primtivies.

The second issue dealt with the challenge of computing at Uber's scale.
After running a few benchmarks, we found that we could not get the standard
distributed TensorFlow to scale as well as our services required. For example,
we lost about half of our resources due to communication overhead when training
on 128 GPUs.

\begin{figure}[h]
  \centering
  \fbox{\includegraphics[width=13cm]{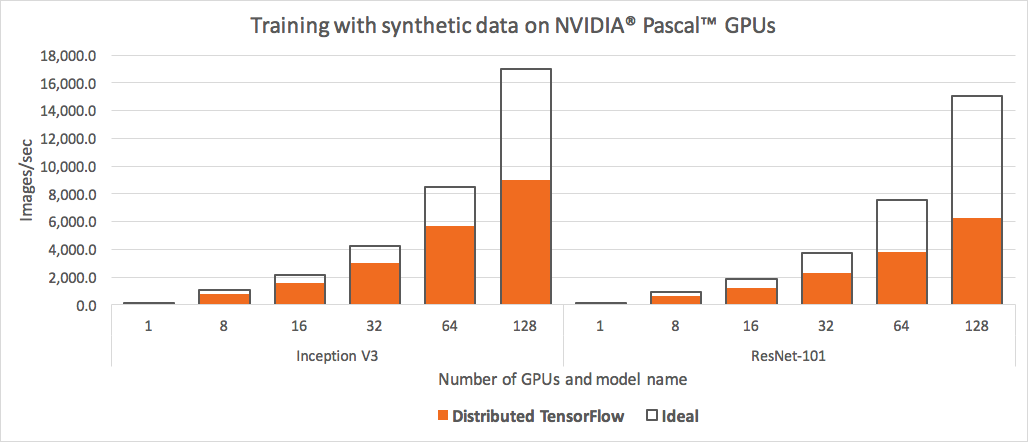}}
  \caption{Multi-GPU scaling performance using TensorFlow. When comparing images
	processed per second while running the standard TensorFlow benchmarking
	suite on NVIDIA Pascal GPUs (ranging from 1 to 128) with both the
	Inception V3 and ResNet-101 TensorFlow models to theoretically ideal
	scaling (computed by multiplying the single-GPU rate by the number of
	GPUs), we were unable to take full advantage of our hardware resources.}
  \label{tensorflow_performance}
\end{figure}

When we ran the standard TensorFlow benchmarking suite~\cite{tensorflow_benchmarks}
on 128 NVIDIA Pascal GPUs, showcased in Figure~\ref{tensorflow_performance}, we
observed that both the Inception V3 and ResNet-101 models were were unable to
leverage nearly half of our GPU resources.

Motivated to make the most of our GPU capacity, we became even more excited about
distributed training after Facebook published a paper~\cite{imagenet_1hr},
demonstrating training of a ResNet-50 network in one hour on 256 GPUs by combining
principles of data parallelism~\cite{data_parallelism} with an innovative learning
rate adjustment technique. This milestone made it abundantly clear that large-scale
distributed training can have an enormous impact on model developer productivity.

\section{Leveraging a different type of algorithm}

\begin{figure}[h]
  \centering
  \fbox{\includegraphics[width=13cm]{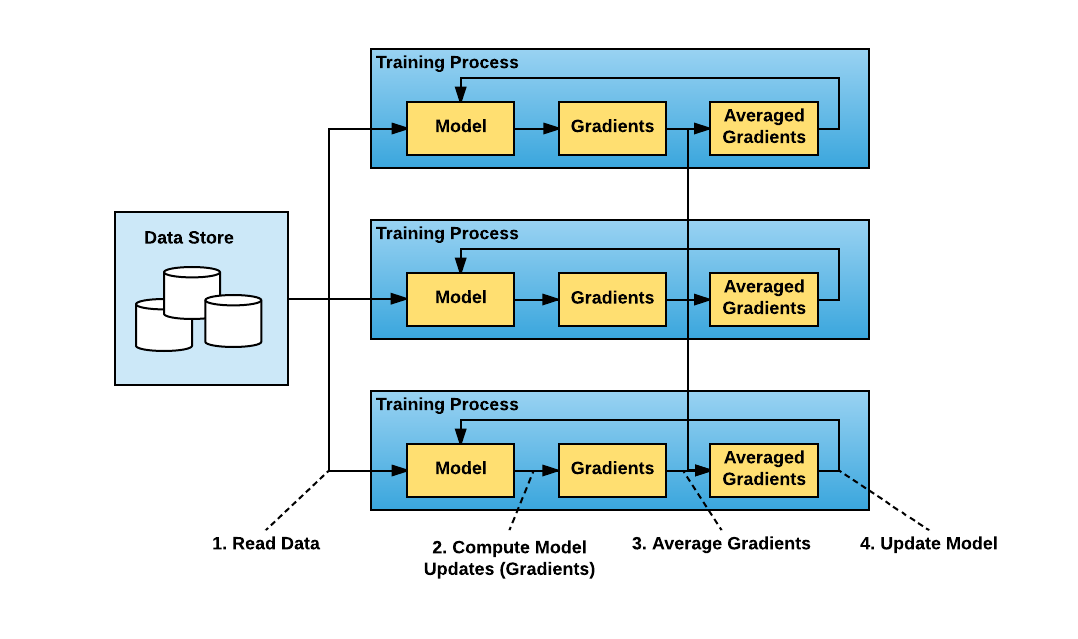}}
  \caption{The “data parallel” approach to distributed training involves splitting up
	the data and training on multiple nodes in parallel. In synchronous cases,
	the gradients for different batches of data are calculated separately on each
	node but averaged across nodes to apply consistent updates to the model copy
	in each node.}
  \label{data_parallel_training}
\end{figure}

After this realization, we started looking for a better way to train our distributed
TensorFlow models. Since our models were small enough to fit on a single GPU, or
multiple GPUs in a single server, we tried using Facebook's data parallel approach
to distributed training, shown on Figure~\ref{data_parallel_training}.

Conceptually, the data-parallel distributed training paradigm is straightforward:

\begin{enumerate}
  \item Run multiple copies of the training script and each copy:
  \begin{enumerate}
    \item \label{itm:repeat} reads a chunk of the data
    \item runs it through the model
    \item computes model updates (gradients)
  \end{enumerate}
  \item \label{itm:average} Average gradients among those multiple copies
  \item \label{itm:update} Update the model
  \item Repeat (from Step~\ref{itm:repeat})
\end{enumerate}

\begin{figure}[h]
  \centering
  \fbox{\includegraphics[width=13cm]{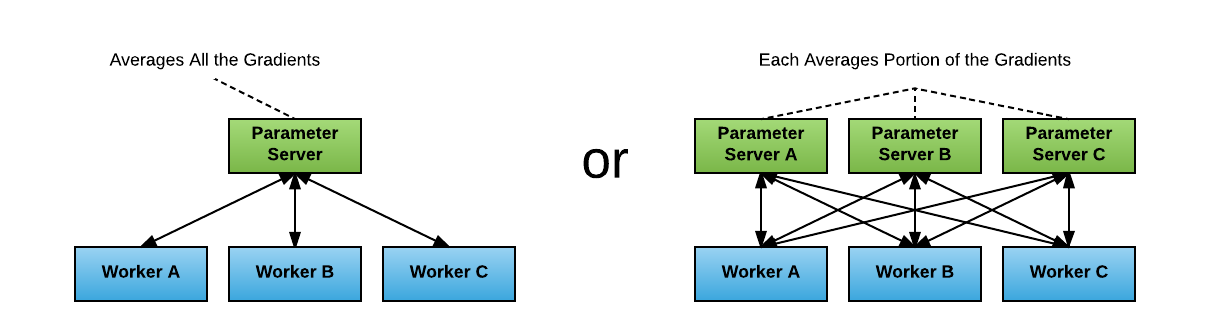}}
  \caption{The parameter server model for distributed training jobs can be configured
	with different ratios of parameter servers to workers, each with different
	performance profiles.}
  \label{parameter_server_architecture}
\end{figure}

The standard distributed TensorFlow package runs with a parameter server approach
to averaging gradients, shown on Figure~\ref{parameter_server_architecture}. In this
approach, each process has one of two potential roles: a worker or a parameter
server. Workers process the training data, compute gradients, and send them to
parameter servers to be averaged.

While this approach improved our performance, we encountered two challenges:

\begin{itemize}
  \item \textbf{Identifying the right ratio of worker to parameter servers:} If one
	parameter server is used, it will likely become a networking or computational
	bottleneck. If multiple parameter servers are used, the communication pattern
	becomes “all-to-all” which may saturate network interconnects.
  \item \textbf{Handling increased TensorFlow program complexity:} During our testing,
	every user of distributed TensorFlow had to explicitly start each worker and
	parameter server, pass around service discovery information such as hosts and
	ports of all the workers and parameter servers, and modify the training
	program to construct \texttt{tf.Server()} with an appropriate \texttt{tf.ClusterSpec()}.
	Additionally, users had to ensure that all the operations were placed
	appropriately using \texttt{tf.train.device\_replica\_setter()} and code is modified
	to use towers to leverage multiple GPUs within the server. This often led to
	a steep learning curve and a significant amount of code restructuring, taking
	time away from the actual modeling.
\end{itemize}

In early 2017 Baidu published an article~\cite{baidu_hpc_techniques} evangelizing
a different algorithm for averaging gradients and communicating those gradients to
all nodes (Steps~\ref{itm:average} and~\ref{itm:update} above), called ring-allreduce,
as well as a fork of TensorFlow through which they demonstrated a draft implementation
of this algorithm. The algorithm was based on the approach introduced in the 2009
paper by Patarasuk and Yuan~\cite{bandwidth_optimal_allreduce}.

\begin{figure}[h]
  \centering
  \fbox{\includegraphics[width=13cm]{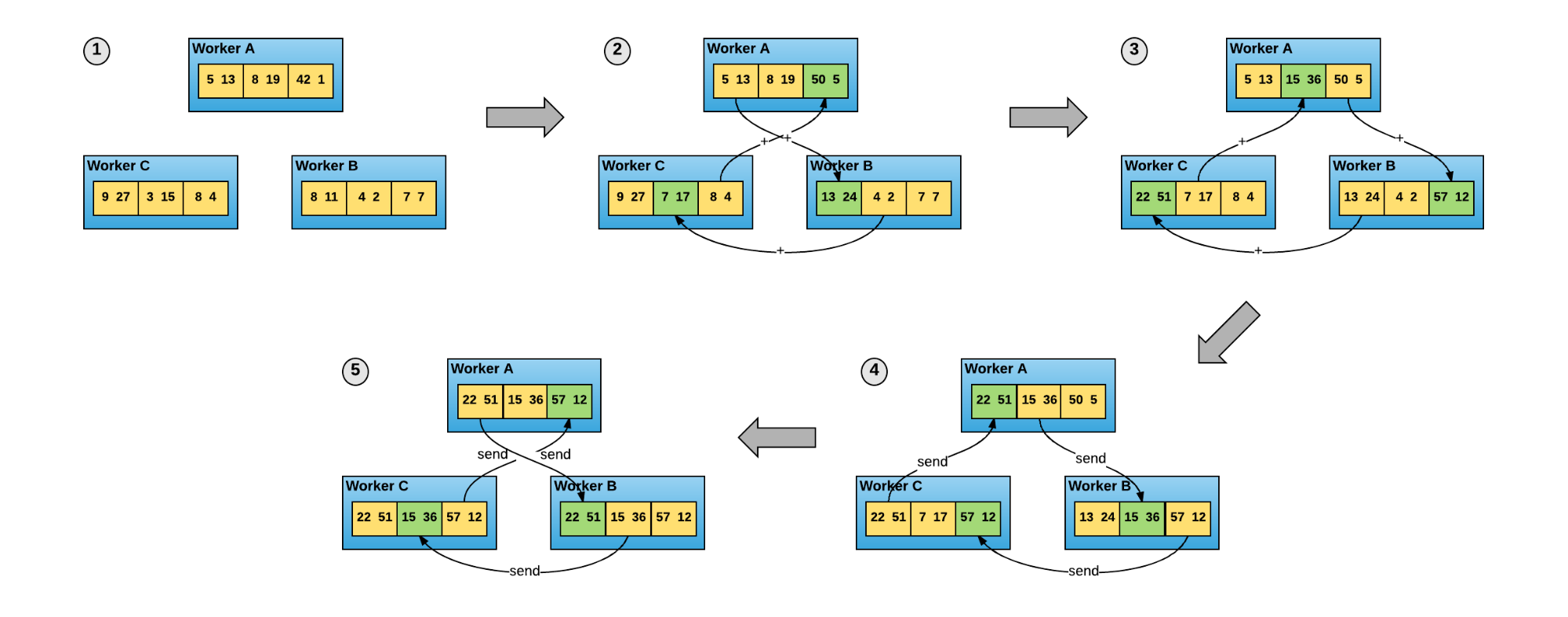}}
  \caption{The ring-allreduce algorithm allows worker nodes to average gradients and
	disperse them to all nodes without the need for a parameter server.}
  \label{ring_allreduce_algorithm}
\end{figure}

In the ring-allreduce algorithm, shown on Figure~\ref{ring_allreduce_algorithm}, each
of \(N\) nodes communicates with two of its peers \(2*(N-1)\) times. During this
communication, a node sends and receives chunks of the data buffer. In the first \(N-1\)
iterations, received values are added to the values in the node's buffer. In the second
\(N-1\) iterations, received values replace the values held in the node's buffer.
Patarasuk and Yuan in~\cite{bandwidth_optimal_allreduce} suggest that this algorithm is
bandwidth-optimal, meaning that if the buffer is large enough, it will optimally utilize
the available network.

In addition to being network-optimal, the allreduce approach is much easier to understand
and adopt. Users utilize a Message Passing Interface (MPI)~\cite{mpi_forum} implementation
such as Open MPI~\cite{openmpi} to launch all copies of the TensorFlow program. MPI then
transparently sets up the distributed infrastructure necessary for workers to communicate
with each other.  All the user needs to do is modify their program to average gradients
using an \texttt{allreduce()} operation.

\section{Introducing Horovod}

The realization that a ring-allreduce approach can improve both usability and performance
motivated us to work on our own implementation to address Uber's TensorFlow needs. We
adopted Baidu's draft implementation~\cite{tensorflow_allreduce} of the TensorFlow ring-allreduce
algorithm and built upon it. We outline our process below:

\begin{enumerate}
  \item We converted the code into a stand-alone Python package called Horovod, named
	after a traditional Russian folk dance in which performers dance with linked
	arms in a circle, much like how distributed TensorFlow processes use Horovod to
	communicate with each other. At any point in time, various teams at Uber may be
	using different releases of TensorFlow. We wanted all teams to be able to leverage
	the ring-allreduce algorithm without needing to upgrade to the latest version of
	TensorFlow, apply patches to their versions, or even spend time building out the
	framework. Having a stand-alone package allowed us to cut the time required to
	install Horovod from about an hour to a few minutes, depending on the hardware.
  \item We replaced the Baidu ring-allreduce implementation with NCCL~\cite{nccl}. NCCL
	is NVIDIA's library for collective communication that provides a highly optimized
	version of ring-allreduce. NCCL 2 introduced the ability to run ring-allreduce
	across multiple machines, enabling us to take advantage of its many performance
	boosting optimizations.
  \item We added support for models that fit inside a single server, potentially on
	multiple GPUs, whereas the original version only supported models that fit on
	a single GPU.
  \item Finally, we made several API improvements inspired by feedback we received from
	a number of initial users. In particular, we implemented a broadcast operation
	that enforces consistent initialization of the model on all workers. The new API
	allowed us to cut down the number of operations a user had to introduce to their
	single GPU program to four.
\end{enumerate}

Next, we discuss how you can use Horovod for your team's machine learning use cases, too!

\section{Distributing your training job with Horovod}

Whereas the parameter server paradigm for distributed TensorFlow training often requires
careful implementation of significant boilerplate code~\cite{boilerplate_code_example},
Horovod needs just a few new lines. In Listing~\ref{horovod_example}, we offer an example
of a TensorFlow program distributed using Horovod.

\begin{longlisting}
\begin{minted}[frame=lines]{python}
import tensorflow as tf
import horovod.tensorflow as hvd

# Initialize Horovod
hvd.init()

# Pin GPU to be used to process local rank (one GPU per process)
config = tf.ConfigProto()
config.gpu_options.visible_device_list = str(hvd.local_rank())

# Build model...
loss = ...
opt = tf.train.AdagradOptimizer(0.01)

# Add Horovod Distributed Optimizer
opt = hvd.DistributedOptimizer(opt)

# Add hook to broadcast variables from rank 0 to all other processes
# during initialization.
hooks = [hvd.BroadcastGlobalVariablesHook(0)]

# Make training operation
train_op = opt.minimize(loss)

# The MonitoredTrainingSession takes care of session initialization,
# restoring from a checkpoint, saving to a checkpoint, and closing
# when done or an error occurs.
with tf.train.MonitoredTrainingSession(checkpoint_dir="/tmp/train_logs",
                                       config=config,
                                       hooks=hooks) as mon_sess:
 while not mon_sess.should_stop():
   # Perform synchronous training.
   mon_sess.run(train_op)
\end{minted}
\caption{Example TensorFlow program distributed using Horovod.}
\label{horovod_example}
\end{longlisting}

As this example shows, there are only a few changes necessary to make
single-GPU programs distributed:

\begin{enumerate}
  \item \texttt{hvd.init()} initializes Horovod.
  \item \texttt{config.gpu\_options.visible\_device\_list = str(hvd.local\_rank())}
	assigns a GPU to each of the TensorFlow processes.
  \item \texttt{opt=hvd.DistributedOptimizer(opt)} wraps any regular TensorFlow optimizer
	with Horovod optimizer which takes care of averaging gradients using ring-allreduce.
  \item \texttt{hvd.BroadcastGlobalVariablesHook(0)} broadcasts variables from the first
	process to all other processes to ensure consistent initialization. If the program
	does not use \texttt{MonitoredTrainingSession}, users can run the
	\texttt{hvd.broadcast\_global\_variables(0)} operations instead.
\end{enumerate}

User can then run several copies of the program across multiple servers using the
\texttt{mpirun} command:

\texttt{\$ mpirun -np 16 -H server1:4,server2:4,server3:4,server4:4 python train.py}

The \texttt{mpirun} command distributes \texttt{train.py} to four nodes and runs it on
four GPUs per node.

Horovod can also distribute Keras programs by following the same steps.  (You can find
examples of scripts for both TensorFlow and Keras on the Horovod GitHub
page~\cite{horovod_github}.)

Horovod's ease of use, debugging efficiency, and speed makes it a highly effective
sidekick for engineers and data scientists interested in distributing a single-GPU or
single-server program. Next, we introduce Horovod Timeline, a means of providing a high
level of understanding of the states of worker nodes during a distributed training job.

\section{Horovod Timeline}

As we onboarded users to Horovod, we realized that we needed a way for them to easily
identify bugs in their code, an issue commonly faced when dealing with complex distributed
systems. In particular, it was difficult to use native TensorFlow timelines or the CUDA
Profiler because users are required to collect and cross-reference profiles from the
various servers.

\begin{figure}[h]
  \centering
  \fbox{\includegraphics[width=13cm]{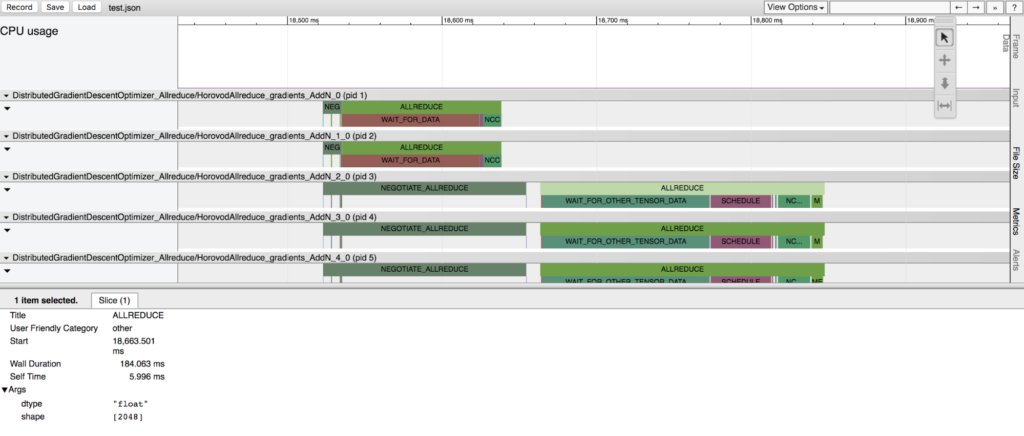}}
  \caption{Horovod Timeline depicts a high level timeline of events in a distributed
	training job in Chrome's trace event profiling tool.}
  \label{horovod_timeline_figure}
\end{figure}

With Horovod, we wanted to created a way to provide a high-level understanding of operation
timelines across nodes. To do so, we built Horovod Timeline, a Horovod-focused profiling
tool compatible with Chrome's \texttt{about:tracing}~\cite{about_tracing} trace event
profiling viewer. Users can use Horovod Timelines to view exactly what each node was doing
at each time step throughout a training job. This helps identify bugs and debug performance
issues. Users can enable timelines by setting a single environment variable and can view
the profiling results in the browser through \texttt{chrome://tracing}.
Figure~\ref{horovod_timeline_figure} shows an example of Horovod Timeline.

\section{Tensor Fusion}

After we analyzed the timelines of a few models, we noticed that those with a large amount
of tensors, such as ResNet-101, tended to have many tiny \texttt{allreduce} operations. As
noted earlier, ring-allreduce utilizes the network in an optimal way if the tensors are
large enough, but does not work as efficiently or quickly if they are very small. We asked
ourselves: what if multiple tiny tensors could be fused together before performing
ring-allreduce on them?

Our answer: Tensor Fusion, an algorithm that fuses tensors together before we call Horovod's
ring-allreduce. As we experimented with this approach, we observed up to 65 percent improvement
in performance on models with a large number of layers running on an unoptimized transmission
control protocol (TCP) network. We outline how to use Tensor Fusion, below:

\begin{enumerate}
  \item Determine which tensors are ready to be reduced. Select the first few tensors that
	fit in the buffer and have the same data type.
  \item Allocate a fusion buffer if it was not previously allocated. Default fusion buffer
	size is 64~MB.
  \item Copy data of selected tensors into the fusion buffer.
  \item Execute the allreduce operation on the fusion buffer.
  \item Copy data from the fusion buffer into the output tensors.
  \item Repeat until there are no more tensors to reduce in the cycle.
\end{enumerate}

With Horovod, Tensor Fusion, and other features built on top of Michelangelo, we can increase
the efficiency, speed, and ease-of-use across our machine learning systems. In our next section,
we share real world benchmarks that showcase Horovod's performance.

\section{Horovod Benchmarks}

\begin{figure}[h]
  \centering
  \fbox{\includegraphics[width=13cm]{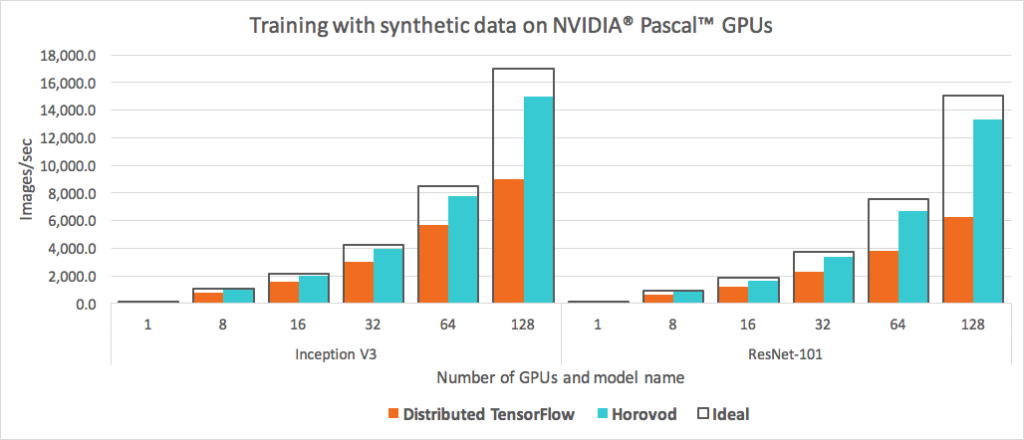}}
  \caption{A comparison of images processed per second with standard distributed TensorFlow
	and Horovod when running a distributed training job over different numbers of NVIDIA
	Pascal GPUs for Inception V3 and ResNet-101 TensorFlow models over 25GbE TCP.}
  \label{horovod_benchmark_tcp}
\end{figure}

We re-ran the official TensorFlow benchmarks modified to use Horovod~\cite{horovod_benchmarks}
and compared the performance with regular distributed TensorFlow. As depicted in
Figure~\ref{horovod_benchmark_tcp}, we observed large improvements in our ability to scale;
we were no longer wasting half of the GPU resources—in fact, scaling using both Inception V3
and ResNet-101 models achieved an 88 percent efficiency mark. In other words, the training was
about twice as fast as standard distributed TensorFlow.

\begin{figure}[h]
  \centering
  \fbox{\includegraphics[width=13cm]{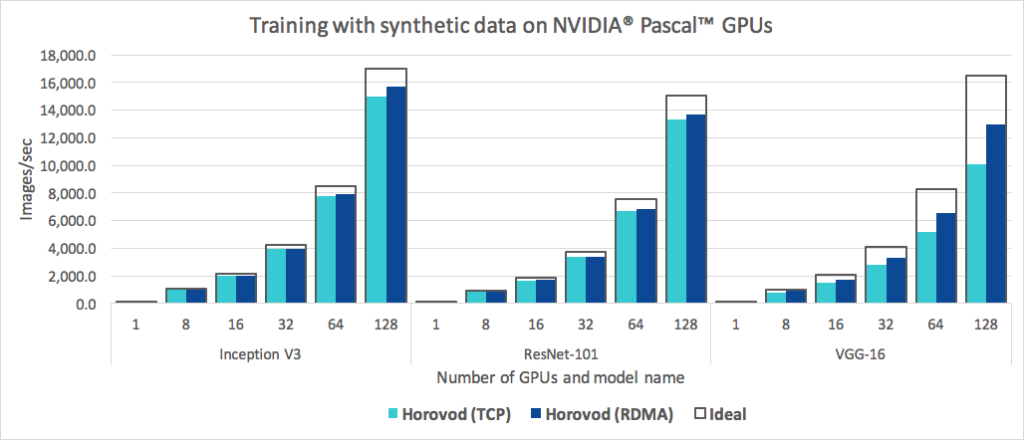}}
  \caption{A comparison of the images processed per second of the Horovod over plain 25GbE TCP
	and the Horovod with 25GbE RDMA-capable networking when running a distributed training
	job over different numbers of NVIDIA Pascal GPUs for Inception V3, ResNet-101 and VGG-16.}
  \label{horovod_benchmark_rdma}
\end{figure}

Since both MPI and NCCL support remote direct memory access (RDMA)~\cite{rdma} capable networking
(e.g., via InfiniBand~\cite{infiniband} or RDMA over Converged Ethernet~\cite{roce}), we ran
additional sets of benchmarking tests using RDMA network cards to determine if they helped us
enhance efficiency compared to TCP networking. Figure~\ref{horovod_benchmark_rdma} shows the
results of this benchmark.

For the Inception V3 and ResNet-101 models, we found that RDMA did not significantly improve
our performance and only achieved a three to four percent increase over TCP networking. RDMA,
however, did help Horovod exceed 90 percent scaling efficiency on both models.

Meanwhile, the VGG-16 model experienced a significant 30 percent speedup when we leveraged RDMA
networking. This can be explained by VGG-16's high number of model parameters, caused by the use
of fully connected layers combined with its small number of layers. These characteristics shifted
the critical path from GPU computation to communication and created a networking bottleneck.

These benchmarks demonstrate that Horovod scales well on both plain TCP and RDMA-capable networks,
although users with RDMA networking will be able to squeeze out optimal performance and experience
a significant efficiency gain when using models with a high number of model parameters, such as
the VGG-16.

With Horovod, we have only scratched the surface when it comes to exploring performance
optimizations in deep learning; in the future, we intend to continue leveraging the open source
community to extract additional performance gains with our machine learning systems and frameworks.

\section{Next steps}

There are a few areas that we are actively working on to improve Horovod, including:

\begin{enumerate}
  \item \textbf{Making it easier to install MPI:} While it is relatively easy to install MPI on
	a workstation, installation of MPI on a cluster typically requires some effort; for
	instance, there are number of workload managers available and different tweaks should be
	made depending on network hardware. We are developing reference designs for running
	Horovod on a cluster; to do so, we hope to work with the MPI community and network hardware
	vendors to develop instructions for installing MPI and relevant drivers.
  \item \textbf{Collecting and sharing learnings about adjusting model parameters for distributed
	deep learning:} Facebook’s paper~\cite{imagenet_1hr} describes the adjustments needed to
	model hyperparameters to achieve the same or greater accuracy in a distributed training
	job compared to training the same model on a single GPU, demonstrating the feasibility of
	training a TensorFlow model on 256 GPUs. We believe this area of deep learning research
	is still in its early stages and hope to collaborate with other teams about approaches to
	further scale deep learning training.
  \item \textbf{Adding examples of very large models:} Horovod currently supports models that fit
	into one server but may span multiple GPUs. We are eager to develop more examples for large
	models spanning multiple GPUs, and encourage others to test Horovod on these types of
	models as well.
\end{enumerate}

We hope the simplicity of Horovod enables others to adopt distributed training and better leverage
their compute resources for deep learning. We welcome feedback and contributions: please report
any issues you encounter, share speed-ups, and send pull requests.

\section*{Acknowledgements}

The authors would like to thank Molly Vorwerck and Jason Yosinski for the help in preparing this paper.

\bibliography{horovod_arxiv}

\end{document}